\pdfoutput=1

\documentclass[11pt]{article}

\usepackage{acl}

\usepackage{times}
\usepackage{latexsym}

\usepackage{linguex}

\usepackage{booktabs}
\usepackage{amssymb}
\usepackage{amsmath}

\usepackage[T1]{fontenc}
\usepackage[utf8]{inputenc}

\usepackage{microtype}

\usepackage{todonotes}
\usepackage[capitalise,noabbrev]{cleveref}

\usepackage{caption}
\usepackage{subcaption}

\newcommand{\tick}{\textregistered\xspace}
\newcommand{\slide}{\textsc{SLIDE}\xspace}
\newcommand{\slidefull}{\textsc{SLIDE}\ensuremath{_f}\xspace}

\title{SLIDE: Reference-free Evaluation for Machine Translation using a Sliding Document Window}

\author{Vikas Raunak \and Matt Post \and Tom Kocmi \\
  Microsoft \\
  Redmond, Washington, USA \\
  \texttt{\{viraunak,mattpost,tom.kocmi\}@microsoft.com}}

\begin{document}
\maketitle

\begin{abstract}

Reference-based metrics that operate at the sentence-level typically outperform quality estimation metrics, which have access only to the source and system output.
This is unsurprising, since references resolve ambiguities that may be present in the source.
In this paper, we investigate whether additional source context can effectively substitute for a reference.
We present a metric named \textbf{\slide} (\underline{\textbf{SLI}}ding \underline{\textbf{D}}ocument \underline{\textbf{E}}valuator), which operates on blocks of sentences. \slide leverages a moving window that slides over each document in the test set, feeding each chunk of sentences into an unmodified, off-the-shelf quality estimation model.
We find that \slide obtains significantly higher pairwise system accuracy than its sentence-level baseline, in some cases even eliminating the gap with reference-base metrics.
This suggests that source context may provide the same information as a human reference in disambiguating source ambiguities. This finding is especially pertinent for reference-free document-level evaluation, wherein SLIDE could provide higher-quality pairwise system assessments while only requiring document boundary annotations.


\end{abstract}

\section{Introduction}
\label{section:intro}

The prevailing approach for neural machine translation metrics is to work at the sentence-level, constructing sequences of contextualized encoder states from the source sentence, a reference translation, and a system output.
The specific mechanics vary by metric, but a general approach, employed by COMET \citep{rei-etal-2020-unbabels}, is to pool these encodings into separate sentence-level embeddings, concatenate them, and feed them into a regressor, which is trained against human annotations.
Quality Estimation (QE) approaches work similarly, but do not have access to a reference translation.

QE metrics typically trail their reference-based counterparts \citep{freitag-etal-2022-results}, for obvious reasons.
The default evaluation setting for QE is at the sentence-level.
But just as there exist many linguistic phenomena that cannot be translated without context, these same phenomena also cannot be properly evaluated in \textit{isolation}.
As an example, consider the following English sentences \emph{with context} and their translations into German.

\ex.
\a. \emph{I need my hat}. Where is \textbf{it}?
\b. \emph{Ich brauche meinen Hut}. Wo ist \textbf{er}?

Reference-based evaluation is aided by the fact that the human translation, presumably produced in context, resolves the \textbf{ambiguity}.
QE approaches (operating at the sentence-level), on the other hand, cannot correctly score this translation.
There are many other document-level phenomena that are also typically captured by references, many of which are subtle and hard to measure \cite{maruf-etal-2019-survey}.

It therefore stands to reason that providing disambiguating context could be a useful extension to (reference-free) QE metrics.
In this work, we are motivated by two related ideas to address this gap: (i) neural metrics often make use of underlying language models trained on wider contexts, which means there is no real impediment to feeding them multiple sentences, and (ii) a sentence's evaluation will differ based on its order in a block of sentences, so it may be helpful to evaluate each sentence in multiple different contexts. We therefore experiment with a \textbf{strided window} approach applied to COMET, whose underlying encoder is InfoXLM \cite{lample-conneau-2019-cross, chi-etal-2021-infoxlm}, trained on wide contexts.
We apply a fixed-width sentence window and slide it across the documents within a test set, accumulating scores of each chunk in normal COMET fashion.
We experiment with various windows and strides, and find that COMET-QE employed in this fashion outperforms its sentence-level reference-based counterparts in many settings.
We conclude that this simple extension to QE might be profitably engaged wherever document boundary annotations are available.

\section{The SLIDE Approach}
\label{section:approach}

\begin{figure}[t]
    \centering
    \includegraphics[width=\columnwidth]{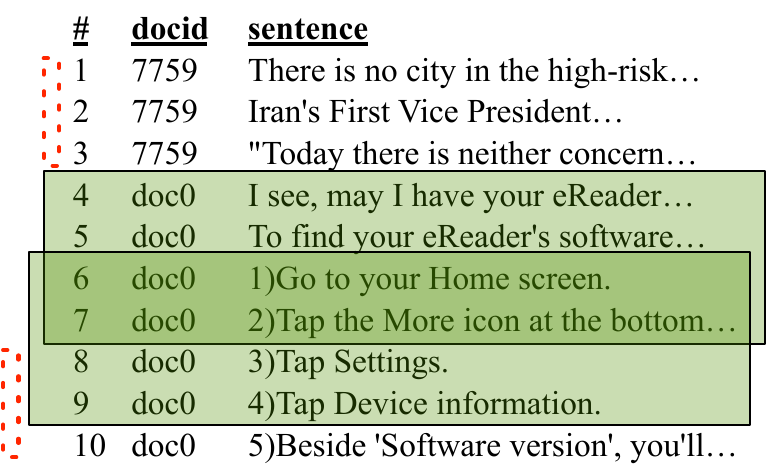}
    \caption{\slide extraction for $(w=4,s=2)$.
    The solid green boxes denote extracted chunks, which are then joined with a space and sent to COMET as a single unit.
    The dashed red boxes denote partial documents: a document that is too short (top), and a document remainder (bottom).}
    \label{figure:slide-example}
\end{figure}

Our main focus is the system ranking question as posed in \citet{kocmi-etal-2021-ship}: given two systems, which one is the better one?

The task is to score system output over a test set, comprising multiple documents.
We define a window size, $w$, specifying how many sentences to include as input, and a stride, $s{\le}w$, defining how many sentences to advance the window.
For a given $(w,s)$ setting, the window is placed at the beginning of the document covering $w$ sentences, and the chunk is sent to COMET as a single input.
The window is then incremented by $s$ sentences and a new value computed.
This proceeds over all documents in a test set.
This is depicted in Figure~\ref{figure:slide-example}.

There are a few edge cases that must be considered, both pertaining to situations where we cannot fill a full window:
\begin{itemize}
    \item \textit{Documents smaller than $w$}.
    Windows are constrained by document boundaries.
    If a document is smaller than the window, we must therefore decide whether to include it.
    \item \textit{Document remainder}.
    If a document length $d$ is not evenly divisible by $w-s$, there will be a document remainder at the end of the document of size $(d\bmod w-s)$.
\end{itemize}
In our initial experiments, we ignore both types of partial documents, i.e., throwing them away, as if they didn't exist in order to evaluate only on chunks with appropriate context available.
We will return to this question in Section~\ref{section:partials} to consider alternatives to throwing away partial documents.

Finally, we accumulate the scores from all chunks in a test set, and return their average as the system-level score. We call our metric SLIDE, for \underline{SLI}ding \underline{D}ocument \underline{E}valuator.


%

\begin{figure*}[t]
    \centering
    \includegraphics[width=\textwidth]{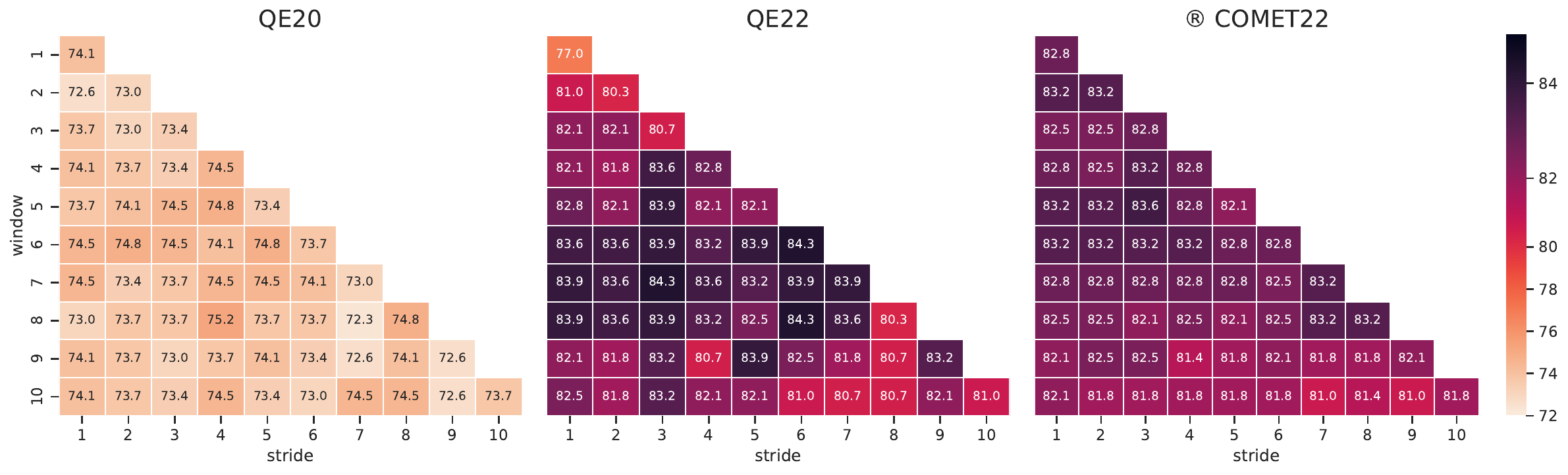}
    \caption{Plot of window vs.\ stride accuracies for QE20, QE22, and {\tick}COMET22 models on the WMT22 MQM task (en-de, en-ru, and zh-en).
    Neither COMET20-QE nor the reference-based {\tick}COMET22 sees much improvement from adding context, but the QE22 model does.}
    \label{figure:heatmap}
\end{figure*}

\section{Experiments}
\label{section:experiments}

\subsection{Evaluation Settings}
\label{section:settings}

Our experiments will explore the performance of various \slide approaches against the WMT22 metric evaluation task \cite{freitag-etal-2022-results}.
These tasks gathered human rankings of system outputs using different methods of collecting those human judgments: MQM and DA+SQM \citep{kocmi-etal-2022-findings}.
The human scores induce a ranking of all system outputs, for each language pair.
The $N$ scored system outputs are collected into a set of $\binom{N}{2}$ pairwise system comparisons across all language pairs, considering pairs from each set of systems per language pair in isolation.
Metrics are then scored based on their ability to correctly predict the better system in each pair.
Table~\ref{table:stats} contains information about the data used in the evaluation.

Our key evaluation metric is the pairwise system-level accuracy \cite{kocmi-etal-2021-ship}, a percentage of pairwise system pairs that a metric correctly distinguishes. We experiment with the following \slide settings:

\begin{table}[t]
    \centering
    \begin{tabular}{c|c|c}
        \toprule
    Judgement Style & Lang-Pairs & System Pairs \\ \midrule
      MQM   & 3  & 274 \\ \midrule
      DA + SQM   &  13 &  564 \\ \bottomrule
    \end{tabular}
    \caption{Data used in the evaluation \cite{freitag-etal-2022-results}}
    \label{table:stats}
\end{table}

\begin{itemize}
    \item Window values $1{\leq}w{\leq}10$ and strides of $1{\leq}s{\leq}w$.
    \item Multiple state-of-the-art QE models (COMET-QE-20 and COMET-QE-22). Further,
    in addition to QE models, we also look at a reference-based metric, COMET22.\footnote{COMET model \texttt{wmt22-comet-da}.}
    \item Incorporating partial documents (\S~\ref{section:partials}).
\end{itemize}

\subsection{Results}

Figure~\ref{figure:heatmap} contains a heatmap depicting pairwise system-level accuracies for each $(w,s)$ value for three models on the MQM task (en-de, en-ru, and zh-en).
The first two are QE20 and QE22, as defined above.
We also include the reference-based {\tick}COMET22\footnote{As a service to the reader, we annotate reference-using models with \tick.} model.\footnote{COMET model \texttt{wmt22-comet-da}}
A number of trends emerge.
The QE20 does not fair well at all with context, with very few points in the grid even improving over the sentence-level baseline (point $(1,1)$) at all, and not in any consistent fashion.
We see the same thing for model {\tick}COMET22, corroborating the finding of \citet{deutsch-etal-2023-training}, as well as our hypothesis that source context may be redundant \textit{with} the reference.

\begin{table}[t]
    \centering
    \begin{tabular}{l|rr}
    \toprule
      \textbf{Metric}
      & \small\textbf{MQM}
      & \small\textbf{DA+SQM}
      \\
    \midrule
    \tick metricx\_xl\_DA\_2019 & 0.865 & 0.850  \\
    \tick metricx\_xxl\_MQM\_2020 & 0.850 & 0.861 \\
    \tick BLEURT-20 &     0.847           & 0.827 \\
    \tick metricx\_xl\_MQM\_2020  &   0.843 & 0.859 \\
    \phantom{\tick}\textbf\slide$(6,6)$ & 0.843 & 0.838\\
    \tick COMET-22 &      0.839  & 0.839 \\
    \phantom{\tick}\textbf\slide$(7,1)$ & 0.839  & 0.814 \\
    \tick COMET-20  &        0.836  & 0.823\\
    \tick Doc-COMET(2) & 	0.836  \\
    \tick UniTE  &     0.828   &  0.847 \\
    \tick MS-COMET-22  &      0.828   & 0.830\\
    \tick UniTE-ref &      0.818  & 0.838 \\
    \tick MATESE   &   0.810  &   \\
    \phantom{\tick}\textbf\slide$(2,1)$ & 0.807 & 0.825\\
    \tick YiSi-1 &  0.792  & 0.782 \\
    \phantom{\tick}COMETKiwi (WMT-22)  &      0.788  & 0.832  \\
    \phantom{\tick}Doc-COMET(2)  & 0.737 & 0.810\\
    \phantom{\tick}COMETKiwi (Public) & 0.770 & 0.816  \\
    \tick chrF &  0.734 & 0.758 \\
    \tick BLEU &  0.708 & 0.704 \\
    \bottomrule
    \end{tabular}
    \caption{Pairwise system accuracy against the WMT22-MQM annotations.
    Metrics that use a reference are marked with \tick.
    Our entries are of the form \slide$(w,s$).
    We have retained many other WMT22 scores for comparison purposes.
    }
    \label{table:results}
\end{table}



With the QE22 model, however, the results show a clear trend.
Adding context helps significantly; the \emph{worst} points in the grid (8-8 and 2-2) improve by 3.3 points over the no-context baseline.
Performance seems to rise as more context is added, although once too much context is used, the points begin to decrease again.
Within a particular window (row), it does not seem to matter too much which stride is used, particularly with window sizes 6--8.
Further, we hypothesize that the purported large gains based on the utilization of source context is not visible in QE20 owing to the lack of cross-attention in its architecture, which strongly suggests that it is not able to leverage the fine-grained source context, which is demonstrably exploited through better modeling of longer-range dependencies in the cross-attention based QE22 model.

Comparing chunked ($w=s$) vs.\ overlapped ($w>s$) values, we also see no particular pattern.
Increasing the context size (up to \textit{some} point) matters, but overlapping the document chunks neither helps nor hurts, on average.

Next, we situate some of these results against leading metrics and other metrics of interest from the WMT22 task in Table~\ref{table:results}.\footnote{We observe that the public model release of COMETKiwi, \texttt{wmt22-cometkiwi-da}, performs notably worse than the one submitted to the WMT22 task.
From private correspondence with the authors, we learned that that the model entered into the shared task was an ensemble.}
We include the best variant of \slide, the worst variant, and, in a nod towards model selection, the best variant with a stride of 1.
Here, we can see that \slide propels quality estimation up the chart, where it even competes with reference-based versions of the underlying evaluators.
In particular, \slide$(6,6)$ outperforms {\tick}COMET22, and a number of variants are at the same level as {\tick}COMET20.
While these results don't answer the difficult question of model selection, the trends empirically validate our key hypothesis (that source-based context could substitute for the reference in evaluation). We note again that even the worst \slide model is already quite an improvement over the baseline.

\subsection{Analysis \& Further Experiments}

We note that it is not obvious that adding context to sentences might improve metric performance when one considers that the machine translation systems being compared are all sentence-based translations.
Any information that is newly available for the evaluator would therefore \emph{not} have been available to the underlying translation engine.
As a result, the improvements the metric is picking up on must come from the new sources included in the evaluation. We further experiment with the QE22 model with different variants of SLIDE in Appendix \ref{appendix_a}. 

\section{Related Work}

The vast majority of work in machine translation metrics is focused on sentence-level evaluation.
At a high level, it is useful to distinguish two uses of context for machine translation evaluation.

In the first setting, contextualized metrics are designed to test document-translation capabilities of a model.
Chief among these are contrastive test sets, in which a model is tested in its ability to rank good translations from bad ones in a contextualized manner \cite{muller-etal-2018-large,bawden-etal-2018-evaluating,voita-etal-2019-good,lopes-etal-2020-document}.
Other approaches have employed test-time NLP tool chains to target and reward correct prediction of discourse phenomena \cite{jiang-etal-2022-blonde,fernandes-etal-2023-translation}.

In the second setting, metrics make use of context to provide or refine their model of what makes a good translation.
This is useful even for ranking sentence-based translation systems.
The first such system may be DocCOMET \cite{vernikos-etal-2022-embarrassingly}, which used context to modify the encodings of sentences, but then removed that context before invoking COMET's classifier.
They looked at both reference-based and QE metrics, and evaluated on Pearson's correlation against human scores.
Context-COMET \cite{hendy-etal-2023-good} took an approach similar to that described here, but was not evaluated at all.
More recently, \citet{deutsch-etal-2023-training} showed that paragraph-level evaluation works just as well for reference-based metrics, even with underlying metrics trained in a sentence-level fashion.
However, they did not experiment with reference-less metrics (the quality estimation task).

Our approach here fits within this second setting.
In contrast to this prior work, our approach requires no changes to the underlying codebase, is evaluated on pairwise system-level accuracy, and focuses on the quality estimation task, where context is the most promising.

\section{Conclusion and Future Work}

Incorporating context into COMET-QE provides critical information that appears to help the metric better adjudicate the difference between systems.
Even just a single sentence of context \textit{drastically} improves the ability of the model to discriminate between systems.
The method works well even though the scores are accumulated over groups of sentences, which is different from the sentence-level manner in which COMET is trained, and even when the accumulated scores come from overlapping blocks.
The results here required no change to the underlying sentence-based evaluators, and in that sense come for free, so long as document boundary annotations are available.
We therefore recommend that source context be included with any neural quality estimation metric.
An interesting further evaluation would throw some document-context translation engines into the mix; we speculate that contextualized quality estimation (like that provided by \slide) should help discriminate those systems too.

\section{Limitations}
In order to leverage the quality-estimation evaluation benefits from SLIDE, document boundaries must be available on the source sentences. Even though this is a benign requirement in most cases, this is a strict limitation on the application of SLIDE in cases where the document boundary annotations are unavailable.

\bibliography{anthology,custom}

\begin{thebibliography}{18}
\expandafter\ifx\csname natexlab\endcsname\relax\def\natexlab#1{#1}\fi

\bibitem[{Bawden et~al.(2018)Bawden, Sennrich, Birch, and Haddow}]{bawden-etal-2018-evaluating}
Rachel Bawden, Rico Sennrich, Alexandra Birch, and Barry Haddow. 2018.
\newblock \href {https://doi.org/10.18653/v1/N18-1118} {Evaluating discourse phenomena in neural machine translation}.
\newblock In \emph{Proceedings of the 2018 Conference of the North {A}merican Chapter of the Association for Computational Linguistics: Human Language Technologies, Volume 1 (Long Papers)}, pages 1304--1313, New Orleans, Louisiana. Association for Computational Linguistics.

\bibitem[{Chi et~al.(2021)Chi, Dong, Wei, Yang, Singhal, Wang, Song, Mao, Huang, and Zhou}]{chi-etal-2021-infoxlm}
Zewen Chi, Li~Dong, Furu Wei, Nan Yang, Saksham Singhal, Wenhui Wang, Xia Song, Xian-Ling Mao, Heyan Huang, and Ming Zhou. 2021.
\newblock \href {https://doi.org/10.18653/v1/2021.naacl-main.280} {{I}nfo{XLM}: An information-theoretic framework for cross-lingual language model pre-training}.
\newblock In \emph{Proceedings of the 2021 Conference of the North American Chapter of the Association for Computational Linguistics: Human Language Technologies}, pages 3576--3588, Online. Association for Computational Linguistics.

\bibitem[{Deutsch et~al.(2023)Deutsch, Juraska, Finkelstein, and Freitag}]{deutsch-etal-2023-training}
Daniel Deutsch, Juraj Juraska, Mara Finkelstein, and Markus Freitag. 2023.
\newblock \href {http://arxiv.org/abs/2308.13506} {Training and meta-evaluating machine translation evaluation metrics at the paragraph level}.

\bibitem[{Fernandes et~al.(2023)Fernandes, Yin, Liu, Martins, and Neubig}]{fernandes-etal-2023-translation}
Patrick Fernandes, Kayo Yin, Emmy Liu, Andr{\'e} Martins, and Graham Neubig. 2023.
\newblock \href {https://doi.org/10.18653/v1/2023.acl-long.36} {When does translation require context? a data-driven, multilingual exploration}.
\newblock In \emph{Proceedings of the 61st Annual Meeting of the Association for Computational Linguistics (Volume 1: Long Papers)}, pages 606--626, Toronto, Canada. Association for Computational Linguistics.

\bibitem[{Freitag et~al.(2022)Freitag, Rei, Mathur, Lo, Stewart, Avramidis, Kocmi, Foster, Lavie, and Martins}]{freitag-etal-2022-results}
Markus Freitag, Ricardo Rei, Nitika Mathur, Chi-kiu Lo, Craig Stewart, Eleftherios Avramidis, Tom Kocmi, George Foster, Alon Lavie, and Andr{\'e} F.~T. Martins. 2022.
\newblock \href {https://aclanthology.org/2022.wmt-1.2} {Results of {WMT}22 metrics shared task: Stop using {BLEU} {--} neural metrics are better and more robust}.
\newblock In \emph{Proceedings of the Seventh Conference on Machine Translation (WMT)}, pages 46--68, Abu Dhabi, United Arab Emirates (Hybrid). Association for Computational Linguistics.

\bibitem[{Goyal et~al.(2021)Goyal, Du, Ott, Anantharaman, and Conneau}]{goyal-etal-2021-larger}
Naman Goyal, Jingfei Du, Myle Ott, Giri Anantharaman, and Alexis Conneau. 2021.
\newblock \href {https://doi.org/10.18653/v1/2021.repl4nlp-1.4} {Larger-scale transformers for multilingual masked language modeling}.
\newblock In \emph{Proceedings of the 6th Workshop on Representation Learning for NLP (RepL4NLP-2021)}, pages 29--33, Online. Association for Computational Linguistics.

\bibitem[{Hendy et~al.(2023)Hendy, Abdelrehim, Sharaf, Raunak, Gabr, Matsushita, Kim, Afify, and Awadalla}]{hendy-etal-2023-good}
Amr Hendy, Mohamed Abdelrehim, Amr Sharaf, Vikas Raunak, Mohamed Gabr, Hitokazu Matsushita, Young~Jin Kim, Mohamed Afify, and Hany~Hassan Awadalla. 2023.
\newblock \href {http://arxiv.org/abs/2302.09210} {How good are gpt models at machine translation? a comprehensive evaluation}.

\bibitem[{Jiang et~al.(2022)Jiang, Liu, Ma, Zhang, Yang, Huang, Sennrich, Cotterell, Sachan, and Zhou}]{jiang-etal-2022-blonde}
Yuchen Jiang, Tianyu Liu, Shuming Ma, Dongdong Zhang, Jian Yang, Haoyang Huang, Rico Sennrich, Ryan Cotterell, Mrinmaya Sachan, and Ming Zhou. 2022.
\newblock \href {https://doi.org/10.18653/v1/2022.naacl-main.111} {{BlonDe}: An automatic evaluation metric for document-level machine translation}.
\newblock In \emph{Proceedings of the 2022 Conference of the North American Chapter of the Association for Computational Linguistics: Human Language Technologies}, pages 1550--1565, Seattle, United States. Association for Computational Linguistics.

\bibitem[{Kocmi et~al.(2022)Kocmi, Bawden, Bojar, Dvorkovich, Federmann, Fishel, Gowda, Graham, Grundkiewicz, Haddow, Knowles, Koehn, Monz, Morishita, Nagata, Nakazawa, Nov{\'a}k, Popel, and Popovi{\'c}}]{kocmi-etal-2022-findings}
Tom Kocmi, Rachel Bawden, Ond{\v{r}}ej Bojar, Anton Dvorkovich, Christian Federmann, Mark Fishel, Thamme Gowda, Yvette Graham, Roman Grundkiewicz, Barry Haddow, Rebecca Knowles, Philipp Koehn, Christof Monz, Makoto Morishita, Masaaki Nagata, Toshiaki Nakazawa, Michal Nov{\'a}k, Martin Popel, and Maja Popovi{\'c}. 2022.
\newblock \href {https://aclanthology.org/2022.wmt-1.1} {Findings of the 2022 conference on machine translation ({WMT}22)}.
\newblock In \emph{Proceedings of the Seventh Conference on Machine Translation (WMT)}, pages 1--45, Abu Dhabi, United Arab Emirates (Hybrid). Association for Computational Linguistics.

\bibitem[{Kocmi et~al.(2021)Kocmi, Federmann, Grundkiewicz, Junczys-Dowmunt, Matsushita, and Menezes}]{kocmi-etal-2021-ship}
Tom Kocmi, Christian Federmann, Roman Grundkiewicz, Marcin Junczys-Dowmunt, Hitokazu Matsushita, and Arul Menezes. 2021.
\newblock \href {https://aclanthology.org/2021.wmt-1.57} {To ship or not to ship: An extensive evaluation of automatic metrics for machine translation}.
\newblock In \emph{Proceedings of the Sixth Conference on Machine Translation}, pages 478--494, Online. Association for Computational Linguistics.

\bibitem[{Lample and Conneau(2019)}]{lample-conneau-2019-cross}
Guillaume Lample and Alexis Conneau. 2019.
\newblock \href {http://arxiv.org/abs/1901.07291} {Cross-lingual language model pretraining}.
\newblock \emph{CoRR}, abs/1901.07291.

\bibitem[{Lopes et~al.(2020)Lopes, Farajian, Bawden, Zhang, and Martins}]{lopes-etal-2020-document}
Ant{\'o}nio Lopes, M.~Amin Farajian, Rachel Bawden, Michael Zhang, and Andr{\'e} F.~T. Martins. 2020.
\newblock \href {https://aclanthology.org/2020.eamt-1.24} {Document-level neural {MT}: A systematic comparison}.
\newblock In \emph{Proceedings of the 22nd Annual Conference of the European Association for Machine Translation}, pages 225--234, Lisboa, Portugal. European Association for Machine Translation.

\bibitem[{Maruf et~al.(2019)Maruf, Saleh, and Haffari}]{maruf-etal-2019-survey}
Sameen Maruf, Fahimeh Saleh, and Gholamreza Haffari. 2019.
\newblock \href {http://arxiv.org/abs/1912.08494} {A survey on document-level machine translation: Methods and evaluation}.
\newblock \emph{CoRR}, abs/1912.08494.

\bibitem[{M{\"u}ller et~al.(2018)M{\"u}ller, Rios, Voita, and Sennrich}]{muller-etal-2018-large}
Mathias M{\"u}ller, Annette Rios, Elena Voita, and Rico Sennrich. 2018.
\newblock \href {https://doi.org/10.18653/v1/W18-6307} {A large-scale test set for the evaluation of context-aware pronoun translation in neural machine translation}.
\newblock In \emph{Proceedings of the Third Conference on Machine Translation: Research Papers}, pages 61--72, Brussels, Belgium. Association for Computational Linguistics.

\bibitem[{Rei et~al.(2020)Rei, Stewart, Farinha, and Lavie}]{rei-etal-2020-unbabels}
Ricardo Rei, Craig Stewart, Ana~C Farinha, and Alon Lavie. 2020.
\newblock \href {https://aclanthology.org/2020.wmt-1.101} {Unbabel{'}s participation in the {WMT}20 metrics shared task}.
\newblock In \emph{Proceedings of the Fifth Conference on Machine Translation}, pages 911--920, Online. Association for Computational Linguistics.

\bibitem[{Rei et~al.(2022)Rei, Treviso, Guerreiro, Zerva, Farinha, Maroti, C.~de Souza, Glushkova, Alves, Coheur, Lavie, and Martins}]{rei-etal-2022-cometkiwi}
Ricardo Rei, Marcos Treviso, Nuno~M. Guerreiro, Chrysoula Zerva, Ana~C Farinha, Christine Maroti, Jos{\'e}~G. C.~de Souza, Taisiya Glushkova, Duarte Alves, Luisa Coheur, Alon Lavie, and Andr{\'e} F.~T. Martins. 2022.
\newblock \href {https://aclanthology.org/2022.wmt-1.60} {{C}omet{K}iwi: {IST}-unbabel 2022 submission for the quality estimation shared task}.
\newblock In \emph{Proceedings of the Seventh Conference on Machine Translation (WMT)}, pages 634--645, Abu Dhabi, United Arab Emirates (Hybrid). Association for Computational Linguistics.

\bibitem[{Vernikos et~al.(2022)Vernikos, Thompson, Mathur, and Federico}]{vernikos-etal-2022-embarrassingly}
Giorgos Vernikos, Brian Thompson, Prashant Mathur, and Marcello Federico. 2022.
\newblock \href {https://aclanthology.org/2022.wmt-1.6} {Embarrassingly easy document-level {MT} metrics: How to convert any pretrained metric into a document-level metric}.
\newblock In \emph{Proceedings of the Seventh Conference on Machine Translation (WMT)}, pages 118--128, Abu Dhabi, United Arab Emirates (Hybrid). Association for Computational Linguistics.

\bibitem[{Voita et~al.(2019)Voita, Sennrich, and Titov}]{voita-etal-2019-good}
Elena Voita, Rico Sennrich, and Ivan Titov. 2019.
\newblock \href {https://doi.org/10.18653/v1/P19-1116} {When a good translation is wrong in context: Context-aware machine translation improves on deixis, ellipsis, and lexical cohesion}.
\newblock In \emph{Proceedings of the 57th Annual Meeting of the Association for Computational Linguistics}, pages 1198--1212, Florence, Italy. Association for Computational Linguistics.

\end{thebibliography}
\bibliographystyle{acl_natbib}

\appendix

\section{Further SLIDE Experiments}
\label{appendix_a}

In order to further test the hypothesis that source-based context could substitute for the reference in evaluation, we experiment with two more variants of the SLIDE approach.

\subsection{Partial windows}
\label{section:partials}

\begin{figure}[ht]
\centering
\includegraphics[width=\columnwidth]{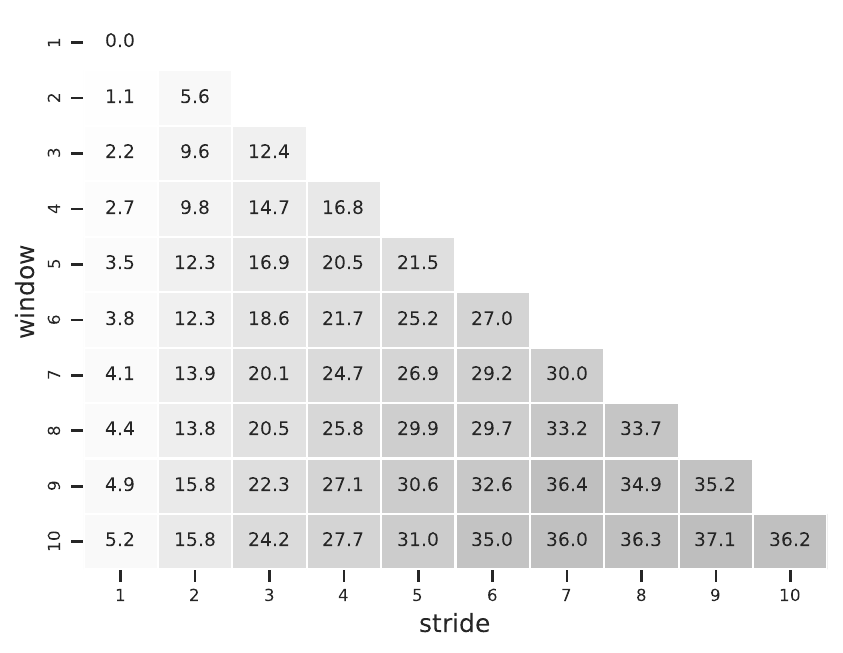}
\caption{Percentage of original-document \emph{sentences} that are dropped when partial documents are ignored.}
\label{figure:dropped}
\end{figure}

The main approach only builds full window sizes.
This means that two types of sentences are excluded: (i) those occurring in test set documents that are shorter than the window size, and (ii) those for which $w-s$ does not evenly divide the document length.
Examples of each of those cases are depicted in \cref{figure:slide-example}. Further, \cref{figure:dropped} lists the percentage of sentences that are left out under the default chunking strategy that omits these two types.

\begin{figure*}
    \centering
    \begin{subfigure}[b]{0.45\textwidth}
        \centering
        \includegraphics[width=\textwidth]{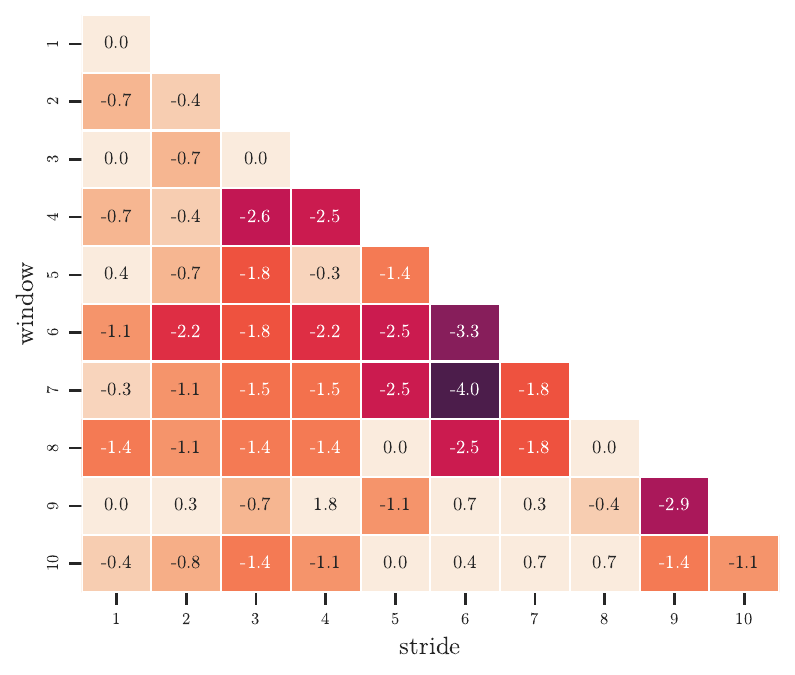}
        \caption{Partial windows.}
        \label{figure:partials:partial}
    \end{subfigure}
    \hfill
    \begin{subfigure}[b]{0.45\textwidth}
        \centering
        \includegraphics[width=\textwidth]{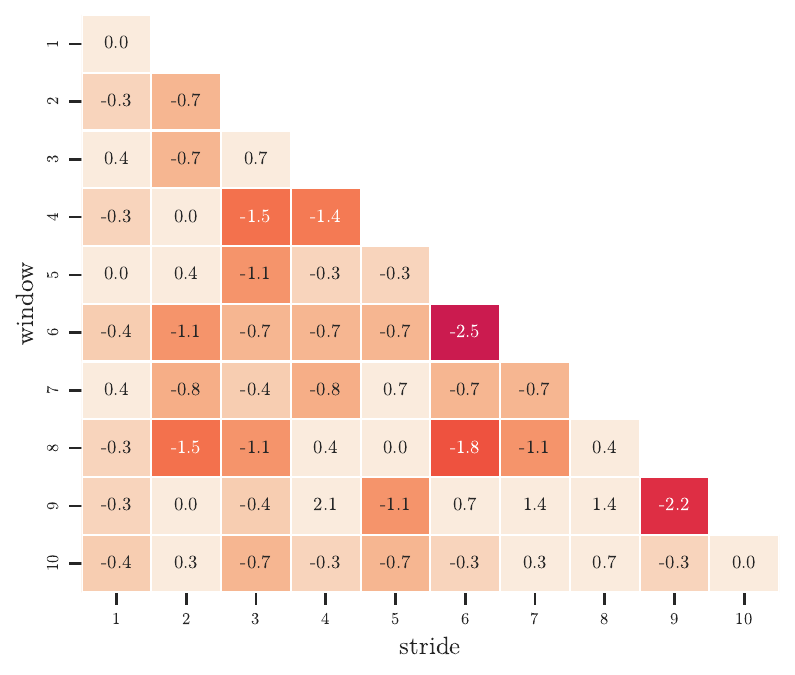}
        \caption{Partial windows with weighting.}
        \label{figure:partials:weighted}
    \end{subfigure}
    \caption{Heatmaps of MQM accuracy difference relative to Figure~\ref{figure:heatmap}(b) for SLIDE when all partials are incorporated.
    In (a), they are treated as equal chunks when producing the document-level score; in (b), the score of each chunk is upweighted based on the number of sentences it contains.
    Both variants improve over context-less QE22, but are generally worse than \slidefull}
    \label{figure:partials}
\end{figure*}

We construct an experiment that includes these partials.
Documents smaller than the window size are included as a complete document, and remainders are added as as partial chunk.
The resulting heatmap for QE22 is depicted in Figure~\ref{figure:partials}~(a).

A problem with blindly incorporating these partials is that they are not properly weighted; in producing the document-level score, all chunks are averaged with a uniform weight.
It stands to reason that smaller chunks should contribute less to the overall score, proportional to their sentence length.
Figure~\ref{figure:partials}~(b) adjusts for this, upweighting each chunk by multiplying its score by the number of sentences it contains, prior to averaging.

Comparing to the middle figure from Figure~\ref{figure:heatmap}, we see a notable drop in system-level accuracy for both plots relative to the only-full-window approach of \slide.
Both systems, however, \textbf{continue to improve} over the original, context-less QE22 model.
We also see that  that weighting the partials helps mitigate the problem.

\subsection{Variable chunking}

\begin{table}[t]
    \centering
    \begin{tabular}{r|rrr}
    \toprule
    $w$ & en-de & en-ru & zh-en \\
    \midrule
    1  & 0     & 0     & 0     \\
    2  & 0     & 0     & 0.2   \\ 
    3  & 0     & 0     & 1.1   \\
    4  & 0.1   & 0.1   & 4.5   \\
    5  & 1.0   & 0.6   & 10.2  \\
    6  & 6.2   & 5.7   & 19.5  \\
    7  & 13.9  & 13.9  & 30.5  \\
    8  & 24.3  & 24.2  & 43.3  \\
    9  & 31.9  & 31.3  & 55.3  \\
    10 & 38.3  & 38.3  & 64.1  \\
    \bottomrule
    \end{tabular}
    \caption{Percentage of \emph{chunks} with a tokenized length $>512$, which means they will be cropped by the underlying InfoXLM model \cite{chi-etal-2021-infoxlm, goyal-etal-2021-larger}.}
    \label{table:clipping}
\end{table}

The problem of partial chunks is not the only design choice when instantiating SLIDE.
Another choice has to deal with managing the maximum token length of the underlying model used by COMET, InfoXLM.
COMET-Kiwi (the QE22 model) encodes the entire input as as single string \cite{rei-etal-2022-cometkiwi}, instead of separately encoding the source and the system output.
This raises the chance that we will hit the maximum token length.
In fact, this is the case.
Table~\ref{table:clipping} lists the percentage of input chunks that rise above the maximum encoding length.

We therefore further experiment with a general solution that will address both the issue of partial document chunks and truncation.
Our approach is to retain the notion of a window and a stride, with sentences as atomic units.
However, the window now specifies a maximum number of tokens.
For a given token-based window size $w_t$, we add (source, system) sentence pairs so long as the number of tokens in the sentence pairs does not exceed $w_t$.
The stride parameter remains purely sentence-based.
To produce the system-level score, we use the sentence-based weighting approach, that multiples each chunk score by the number of sentences in that chunk, prior to averaging.

Unfortunately, this token based chunking approach failed to produce a consistent gain.
Setting a maximum window size of 500 and greedily building chunks as large as possible produced an MQM score of 0.818, considerably lower than the SLIDE results in Table \ref{table:results}.


\end{document}